\documentclass[conference]{IEEEtran}

\usepackage{amsmath,amssymb,amsfonts}
\usepackage{algorithmic}
\usepackage{graphicx}
\usepackage{textcomp}
\usepackage{xcolor}
\usepackage[numbers]{natbib} 

\usepackage[acronym]{glossaries}

\usepackage{csquotes}

\usepackage{pbalance}

\usepackage{listings}
\lstdefinestyle{listingstyle}{
  basicstyle=\ttfamily\footnotesize,
  breaklines=true,
  frame=lines,
  numbers=none,
  captionpos=b,
  keepspaces=true,
}
\usepackage[table]{xcolor}
\definecolor{light-gray}{gray}{0.95}

\usepackage{makecell}

\usepackage{multirow}

\usepackage{mdframed}
\usepackage[breakable, many]{tcolorbox}

\newtcolorbox{answerbox}{
  breakable,
  colback = sub,
  boxrule = 0pt,
  leftrule = 0pt,
}

\usepackage{xurl}

\usepackage{orcidlink}

\hypersetup{
  colorlinks=true,
  linkcolor=black,
  citecolor=black,
  urlcolor=black
}

\usepackage{tabularx} 
\usepackage{booktabs} 

\newcolumntype{C}[1]{>{\centering\arraybackslash}p{#1}} 

\newacronym{ai}{AI}{artificial intelligence}
\newacronym{api}{API}{application programming interface}
\newacronym{dsr}{DSR}{design science research}
\newacronym{hpc}{HPC}{high-performance computing}
\newacronym{llm}{LLM}{large language model}
\newacronym{mcp}{MCP}{Model Context Protocol}
\newacronym{rag}{RAG}{retrieval-augmented generation}

\begin{document}
\bstctlcite{BSTcontrol} 

\title{From Overload to Insights: How {AI} Agents Can Support Scientists in Analyzing Complex Data}

\author{\IEEEauthorblockN{Tim Fuchs \orcidlink{0000-0001-8421-8071}}
  \IEEEauthorblockA{\textit{University of Hamburg} \\
    Hamburg, Germany \\
  tim.fuchs-2@uni-hamburg.de}
  \and
  \IEEEauthorblockN{Luca Gelisio \orcidlink{0000-0001-7832-6201}}
  \IEEEauthorblockA{\textit{European XFEL} \\
    Schenefeld, Germany \\
  luca.gelisio@xfel.eu}
  \and
  \IEEEauthorblockN{Steffen Hauf \orcidlink{0009-0004-1557-694X}}
  \IEEEauthorblockA{\textit{European XFEL} \\
    Schenefeld, Germany \\
  steffen.hauf@xfel.eu}
  \and
  \IEEEauthorblockN{Walid Maalej \orcidlink{0000-0002-6899-4393}}
  \IEEEauthorblockA{\textit{Hasso Plattner Institute} \\
    \textit{University of Potsdam}\\
    Potsdam, Germany \\
    \textit{University of Hamburg}\\
    Hamburg, Germany \\
  walid.maalej@hpi.de}
}

\maketitle

\begin{abstract}
  Scientists at European XFEL conduct experiments that generate very large and complex datasets.
  The subsequent data analysis is challenging as scientists must combine their domain expertise with facility- and software-specific knowledge scattered across documentation, tools, and support channels.
  To address this problem, we designed and evaluated an agentic \gls{ai} system tailored to the scientists' needs and integrated with the high-performance computing environment of European XFEL.
  Using a design science research approach, we conducted a rapid literature review, a systematic evaluation of 16 \gls{ai} tools, multiple interviews, a focus group, and a user study with experts at European XFEL to develop and evaluate two prototypes.
  Our study identifies key knowledge challenges in scientific data analysis, derives requirements for an \gls{ai} agent that supports knowledge retrieval and source code generation, and proposes design recommendations for a specialized system adaptable to the evolving \gls{ai} tool landscape.
  These findings provide guidance for developing maintainable \gls{ai} support in highly specialized scientific environments.
\end{abstract}


\section{Introduction}
\label{sec:intro}

European XFEL is a \textit{user facility} where visiting scientists (\textit{users}) exploit unique types of X-ray pulses to study matter on multiple, complex instruments~\cite{decking_mhz-repetition-rate_2020}.
These users bring strong domain expertise, but they often need help navigating the facility-specific analysis environment and translating experimental data into scientific findings.

The experiments produce extremely large and complex datasets, some on the petabyte scale~\cite{malka_data_2023, sobolev_data_2024}.
To turn this data into scientific results, users need significant computing resources and ad-hoc data analysis software that fit the experimental context.
Like similar facilities, European XFEL enables this process through a \gls{hpc} cluster, specialized software, and user support provided by staff with expertise in photon science, data science, and software engineering.

The user support is distributed across extensive documentation and direct assistance, which makes it strong and comprehensive.
However, this support does not scale well, as the level of direct assistance is constrained by available human resources, while static documentation lacks the contextual guidance required for the broad range of concrete analysis tasks.
These limitations are further amplified by the diversity of users' data analysis experience and needs, and even more so considering that \textit{offline data analysis} can span several months following data collection~\cite{schmidt_turning_2024}.

To improve the current situation, users need a context-sensitive, always available system that bridges the gap between general knowledge resources and their individual workflows~\cite{maalej_task-first_2009}.
\Gls{ai}-based support systems can narrow this divide by turning the existing bulk of documentation into contextualized assistance.
\Gls{rag} is an especially relevant concept because it helps the assistant to ground responses in curated knowledge and reduce \textit{hallucinations}~\cite{james_retrieval-augmented_2025}.
\Gls{ai} agents extend this support by not only answering questions but also planning and executing tasks, including code generation~\cite{sapkota_ai_2026}.
However, for scientific workflows, an agent is only useful if the system behavior remains traceable, controllable, and compatible with human oversight~\cite{tang_risks_2025}.

In this work, we designed and evaluated an agentic \gls{ai} system tailored to offline data analysis at European XFEL.
The system is intended to help users retrieve documentation and generate analysis code while being integrated with the \gls{hpc} cluster.
Our aim was to provide a powerful assistant that is always available and context-sensitive to augment the knowledge resources of the facility~\cite{maalej_task-first_2009, maalej_lightweight_2008}.

We followed a \gls{dsr} approach with seven activities to answer the following research questions:

\begin{description}
  \item[RQ1:] What knowledge challenges do facility users and members encounter during offline data analysis?
  \item[RQ2:] What requirements must an agentic \gls{ai} system meet to address the challenges?
  \item[RQ3:] What design recommendations for effective and maintainable \gls{ai} support can be derived from deploying and evaluating the system in practice?
\end{description}

The remainder of the paper presents the methodology and findings, the discussion, the limitations, and related work.

\section{Design Science Research Approach}
\label{sec:meth}

\subsection{Overview}

\begin{table*}
  \centering
  \caption{Research activities aligned with the research questions and the design science research approach by \citet{peffers_design_2007}}
  \begin{tabularx}{\linewidth}{C{0.15cm} X C{0.55cm} C{0.5cm} C{0.5cm} C{0.5cm} C{0.5cm}}
    \toprule
    & & \textbf{RQ1} & \textbf{RQ2} & \multicolumn{3}{c}{\textbf{RQ3}} \\
    \cmidrule(lr){3-3} \cmidrule(lr){4-4} \cmidrule(lr){5-7}
    \textbf{No.} & \textbf{Activity} & \textbf{PI} & \textbf{OS} & \textbf{DD} & \textbf{DE} & \textbf{EV} \\
    \midrule
    1 & Interviews with 2 managers to discuss the study scope and initial tool concept. & X & X & & & \\
    2 & Focus group with 16 Data Analysis members to analyze workflow and challenges in offline data analysis. & X & & & & \\
    3 & Interviews with 3 facility users and 5 members to analyze offline data analysis challenges and requirements for an \gls{ai} agent. Debriefing with 13 Data Analysis members. & X & X & & & \\
    4 & Design/demonstration first prototype. Evaluation with 12 Data Analysis members. Requirements reassessment. & & X & X & X & X \\
    5 & Rapid multivocal review of 224 articles to integrate insights from related studies into the requirements. & & X & & & \\
    6 & Evaluation of solutions for the agentic AI system components. Reassessment of requirements. & & X & X & & \\
    7 & Design/demonstration of second prototype. Evaluation with 13 experts. Iterative prototype improvement. & & X & X & X & X \\
    \bottomrule
  \end{tabularx} 

  \vspace{2pt}
  \footnotesize{\textit{PI = problem identification and motivation, OS = definition of the objectives for a solution, DD = design and development, DE = demonstration,\\EV = evaluation, Step 6 = communication (not listed above; we communicate our findings in this paper)}}
  \label{tab:meth}
\end{table*}

\Gls{dsr} is appropriate for this exploratory study because it focuses on designing and evaluating IT artifacts for organizational problems \cite{peffers_design_2007}.
This study proceeds through iterative cycles of problem identification, requirements engineering, tool design, and evaluation.
\Gls{dsr} makes the methodological logic of these activities explicit \cite{engstrom_how_2020}.

\autoref{tab:meth} maps our research activities to the six \gls{dsr} steps according to \citet{peffers_design_2007}, as well as to our research questions.
We conducted seven activities, each aligned with at least one \gls{dsr} step.
The process was iterative: the insights gained from the first prototype informed the requirements analysis, design, and evaluation of the second prototype (\autoref{fig:prototype}).
The following subsections describe the activities in chronological order.
Supporting material, such as interview minutes, detailed software evaluation artifacts, and the prototype setup, is available in the \textbf{replication package}~\cite{fuchs_reppack_2026}.

\subsection{Initial Interviews with Managers}

The first author conducted five semi-structured interviews with the two managers of the Data Analysis and Controls groups at European XFEL between June and December 2025.
The objectives were to iteratively scope the challenges during offline data analysis and to identify initial design objectives for a solution.
We selected both managers as participants of this exploratory first activity because their roles place them close to the software, infrastructure, and user support for data analysis, as well as to \gls{ai} initiatives at the facility.

The interviews showed that support requests of users concentrate on three phases: the early phase of experiment preparation, the experiment execution phase, including online (real-time) data analysis, and the late phase of offline data analysis.
In all phases, users must combine their scientific domain expertise with facility-specific knowledge about data formats, instrument settings, and the \gls{hpc} environment.
The managers also emphasized that the existing software manuals of the facility are comprehensive but may be too broad to guide concrete analysis tasks effectively.
This is why the Data Analysis group started to move toward more experiment- and technique-oriented documentation recently~\cite{euxfel_fxe_2026}.

When the discussion turned to solution objectives, the managers converged on the need for first-level support during offline data analysis.
They saw value in a system that can retrieve relevant documentation, provide tailored Jupyter notebook templates, help users navigate complex datasets, optimize analysis code for the \gls{hpc} environment, and report agent conversations to staff to improve traceability.
These capabilities were seen as especially relevant outside normal support hours, when users still need guidance but staff availability is limited.

The interviews also left one important design question unresolved: whether the system should primarily support beginners, experienced users, or both.
We therefore treated these initial interviews as scoping activities and used the results to inform the subsequent focus group and interviews with facility users and members, where we refined the requirements and narrowed the prototype design.

\subsection{Focus Group with Data Analysis Group}

Our objectives for the focus group were to characterize the current offline data analysis workflow at European XFEL and to identify the knowledge-related challenges within this process.
We chose a focus group because the participants shared the same domain and could build upon one another's experience~\cite{kontio_focus_2008}.
The 60-minute session brought together 16 members of the Data Analysis group in January 2026 and was moderated by the first author.
We used a digital mind map to take notes transparently~\cite{luke_improving_2014} and shared the mind map with participants after the session to support member checking~\cite{burgessallen_using_2010}.

The discussion showed that offline data analysis is highly variable across scientific instruments and experiments, but it still follows a recurring pattern.
In broad terms, the work starts with data acquisition and data transfer to the offline storage of the \gls{hpc} environment, continues with the actual data analysis, and ends with scientific findings.
During these steps, users inspect experiment logs~\cite{maia_mylog_2025, maia_mymdc_2025}, exchange with staff, set up an analysis environment, and develop analysis scripts based on templates, libraries, and other software relevant to the experiment.
A typical setup consists of a JupyterLab server on the \gls{hpc} cluster~\cite{reppin_interactive_2021}, a software module with multiple libraries prepared by the Data Analysis group, and further user-provided software.

A second finding was that relevant knowledge is fragmented across many sources.
Participants pointed to general user documentation of the Data Analysis group, library documentation, instrument-specific resources, technical reports, publications (e.g., by~\citet{turkot_extra-xwiz_2023}), experiment proposals, and documentation of the \gls{hpc} cluster.
They emphasized that the main challenge is not the existence of these different resources, but the need to combine them into a workflow for a concrete analysis objective.
This is the point at which an \gls{ai} system could add significant value by retrieving and synthesizing relevant documentation rather than merely listing it~\cite{james_retrieval-augmented_2025}.

\begin{table}
  \centering
  \caption{(RQ1) challenges in offline data analysis at European XFEL}
  \begin{tabularx}{\linewidth}{C{0.2cm} X}
    \toprule
    \textbf{No.} & \textbf{Challenges} \\
    \midrule
    & \textbf{Users:} \\
    1 & Clarity of data analysis objectives \\
    2 & Mismatch between expected and actual data analysis complexity \\
    3 & Limited facility support during offline data analysis phase \\
    4 & Unfamiliarity with infrastructure complexity \\
    5 & Unfamiliarity with high-performance computing \\
    6 & Unfamiliarity with data analysis procedures of instruments \\
    7 & Retrieval and synthesis of scattered documentation \\
    \midrule
    & \textbf{Members of Data Analysis group:} \\
    8 & Unfamiliarity with data analysis procedures of instruments \\
    9 & Retrieval and synthesis of scattered documentation \\
    10 & Understanding of analysis objectives of users \\
    11 & Time effort for user support \\
    \midrule
    & \textbf{Staff scientists:} \\
    12 & Understanding of analysis objectives of users \\
    13 & Time effort for user support \\
    \bottomrule
  \end{tabularx}
  \label{tab:chal}
\end{table}

Overall, the focus group revealed three overlapping sets of 13 challenges, which are summarized in \autoref{tab:chal}.
Users may struggle with concretizing their analysis objectives and steps, understanding the infrastructure, \gls{hpc} environment, and analysis procedures, and synthesizing the scattered documentation.
Members of the Data Analysis group face the same documentation challenge, but they also need to infer users' analysis objectives and spend time supporting them.
Staff scientists experience a similar support burden, especially when objectives are unclear.

The findings established the problem space related to the first research question and informed the subsequent interviews with facility users and members, where we refined the challenge set and translated it into requirements.

\subsection{Interviews with Facility Users and Members}

We conducted informal interviews during the poster session of the annual European XFEL and DESY Users' Meeting in January 2026.
Our objective was to capture firsthand experiences with knowledge challenges in offline data analysis and to explore expectations for an \gls{ai} assistant.
The first author spoke with three facility users and five members, including four staff scientists and one manager.
All participants had experience with offline data analysis and the available user support.
Despite the limited participation, the interview notes and the subsequent debriefing with 13 members of the Data Analysis group helped to refine the insights regarding problem identification and solution objectives.

Two members emphasized that staff scientists and users highly value all support from the Data Analysis group.
However, one user noted that reading documentation is tedious.
This comment reinforces the challenge of retrieving and synthesizing scattered documentation into a usable workflow, which the earlier focus group had already identified.

The interviewees were most engaged when discussing possible features of an \gls{ai} assistant for data analysis.
The features they considered most valuable were support for technical questions about facility software and infrastructure, plotting complex data, and optimizing code for \gls{hpc} execution.
They also emphasized that the system must not send data to third parties.
More optional feature suggestions included adapting the style of generated code to user preferences and providing daily summaries of experiment logs.
Finally, during the debriefing, the Data Analysis group identified a new risk due to \gls{ai} assistance: an agentic system could increase the demand for human support if users begin reporting issues with \gls{ai}-generated code.

These interviews broadened our analysis by adding views from users and staff scientists.
They confirmed the challenges mentioned by the Data Analysis group and provided design objectives for the first prototype, which we designed next.

\subsection{Design, Demonstration, and Evaluation of First Prototype}

We designed and evaluated the first prototype in early March 2026 as a proof of concept for first-level user support during offline data analysis.
To align the prototype with the typical analysis setup, we selected Jupyter AI as the user interface, which integrates directly into Jupyter Lab.
The prototype used the built-in chat persona of Jupyter AI and was connected to a \gls{rag} backend, which was internally developed at European XFEL.
This architecture enabled responses grounded exclusively in curated documentation.

We deployed this system on a local computer and connected Jupyter Lab to a remote Jupyter kernel exposed by the \gls{hpc} cluster.
This configuration gave the \gls{ai} agent access to the data, software, and computational resources needed for offline analysis.
We used GPT-5.4-mini as the chat completion model and text-embedding-3-large for document embedding and knowledge retrieval.
At the time, both were current OpenAI models.
We limited the knowledge base to the documentation repository of EXtra-data, the European XFEL library for reading datasets stored on the \gls{hpc} cluster.
This setup was sufficient for a first demonstration of the solution concept.

We evaluated the prototype in a dedicated meeting with 12 members of the Data Analysis group.
The first author instructed the system to generate a Jupyter notebook for offline data analysis.
The resulting notebook contained code to import EXtra-data, load a dataset, and produce a data visualization.
Participants perceived Jupyter AI as a straightforward user interface, indicating that the familiar notebook environment can lower the barrier to \gls{ai} adoption~\cite{mcnutt_design_2023}.

The evaluation also exposed limitations of the prototype.
The system lacked a sophisticated agent loop to test and repair code autonomously~\cite{sapkota_ai_2026}.
It generated code that we could not fully execute because it contained incorrect library calls.
The participants hypothesized that these \textit{hallucinations} stemmed from a discrepancy between our instructions and the information available in the knowledge base.
Had we granted the system access to the EXtra-data GitHub repository, it might have retrieved additional documentation relevant to our request.
These observations led us to conclude that the next prototype iteration needed better context selection and automatic code verification.
Furthermore, it should support explicit human approval for code execution on the \gls{hpc} cluster to prevent risky autonomous behavior by the agent.

Overall, the first prototype showed that a familiar notebook-based interface is a plausible entry point for \gls{ai} support during offline data analysis.
However, knowledge retrieval alone is not sufficient for reliable code generation.
This feedback motivated a more comprehensive requirements analysis, which we began with a literature review.

\subsection{Rapid, Multivocal Literature Review}

We conducted a rapid, multivocal literature review to extend our empirically derived requirements with further evidence from scientific and practitioner sources.
Because the study aims to inform a practical prototype rather than to produce an exhaustive literature review, we included second- and third-tier grey literature in addition to peer-reviewed publications and followed established guidelines for systematic, rapid, and grey literature reviews~\cite{kitchenham_guidelines_2007, ganann_expediting_2010, garousi_guidelines_2019, tugwell_prisma_2021}.
Our review question focused on collecting requirements for agentic \gls{ai} systems that support scientists in creating source code for data analysis.

We searched Google, Google Scholar, and Consensus AI, and supplemented the search with publications already known to us.
The search terms targeted human-\gls{ai} collaboration, developer experience, and \gls{ai} for science.
We limited the publication window to January 2023 through March 2026 to focus on \gls{llm}-based agents.
The process yielded 224 relevant documents, including 141 peer-reviewed articles, 46 preprints, 36 blog posts, and 1 report.

To accelerate the thematic analysis, we used the \glspl{llm} GPT-5.4-mini and MiniMax-M2.5 as judges and annotators~\cite{wagner_towards_2025}.
The two models annotated 3062 and 2303 candidate requirements, respectively.
Subsequently, GPT-5.4-mini synthesized these candidates into 26 themes.
We manually validated and refined these results into 17 themes, clustered into requirements for system architecture, general agent behavior, knowledge retrieval, code generation, and optional features.

The main trade-off of a rapid review is that it favors speed over exhaustive coverage~\cite{ganann_expediting_2010}.
We did not perform snowballing, and the search strategy was intentionally narrow to align the review with the schedule for the second prototype.
In addition, \glspl{llm} can lead to variability in the results~\cite{ernst_genai_2026}.
We reduced this risk by using two models of different providers, requiring each extracted requirement to be paired with a text quote, and manually validating and reworking the results.

\begin{table}
  \centering
  \caption{(RQ2) requirements for the agentic AI system}
  \begin{tabularx}{\linewidth}{C{0.2cm} X}
    \toprule
    \textbf{No.} & \textbf{The agentic AI system shall ...} \\
    \midrule
    & \textbf{System architecture:} \\
    1 & Access the high-performance computing cluster \\
    2 & Optimize latency during request processing \cite{yamada_ai_2025, schmidgall_agentrxiv_2025} \\
    3 & Not share user data with third parties \cite{bano_qualitative_2025} \\
    4 & Require minimal implementation effort \\
    5 & Require minimal effort to replace components \\
    \midrule
    & \textbf{General agent behavior:} \\
    6 & Interact with content of computational notebook \cite{ramasamy_ai_2025} \\
    7 & Decompose a request into verifiable goals \cite{gu_how_2024,tufano_autodev_2024} \\
    8 & Request human approval for sensitive actions \cite{tang_risks_2025,tufano_autodev_2024} \\
    9 & Close task with narrative summary and recommended steps \cite{fluker_preparing_2025} \\
    10 & Use concise language \\
    \midrule
    & \textbf{Knowledge retrieval:} \\
    11 & Retrieve up-to-date information \cite{wang_code_2025, wehr_virtuous_2025} \\
    12 & Access relevant context information within a project \cite{gao_empowering_2024, wang_code_2025} \\
    13 & Provide and verify citations \cite{olivares_large_2025, wang_scientific_2023} \\
    \midrule
    & \textbf{Code generation:} \\
    14 & Generate code optimized for high-performance computing \cite{dong_survey_2025, rasheed_ai-powered_2024} \\
    15 & Test code for correctness and safety risks \cite{ren_towards_2025, fakhoury_exploring_2024} \\
    16 & Request user feedback to guide and improve solutions \cite{tufano_autodev_2024, herrera_co-explainers_2026} \\
    17 & Adapt code style and explanations to the user's expectations \cite{pudari_copilot_2023, corso_generating_2024} \\
    18 & Generate code documentation on project- and code-level \cite{nejjar_llms_2025,kruse_can_2024} \\
    19 & Recommend using a version control system \cite{kazemitabaar_improving_2024} \\
    \midrule
    & \textbf{Optional requirements:} \\
    20 & Report conversations between user and agent to staff \cite{tufano_autodev_2024, samadi_ai_2024} \\
    21 & Assist in drafting scientific manuscripts \cite{yamada_ai_2025, fuente-ballesteros_artificial_2025} \\
    \bottomrule
  \end{tabularx}
  \label{tab:req}
\end{table}

\autoref{tab:req} lists all requirements that the proposed agentic \gls{ai} system for offline data analysis must fulfill.
Requirements [2-3], [6-9], and [11-21] are (refined) results of the literature review.
The results were sufficient to guide the subsequent evaluation of tool solutions for the second prototype.

\subsection{Evaluation of Solutions for Agentic AI System Components}

\begin{figure*}[t]
  \centering
  \includegraphics[width=\textwidth]{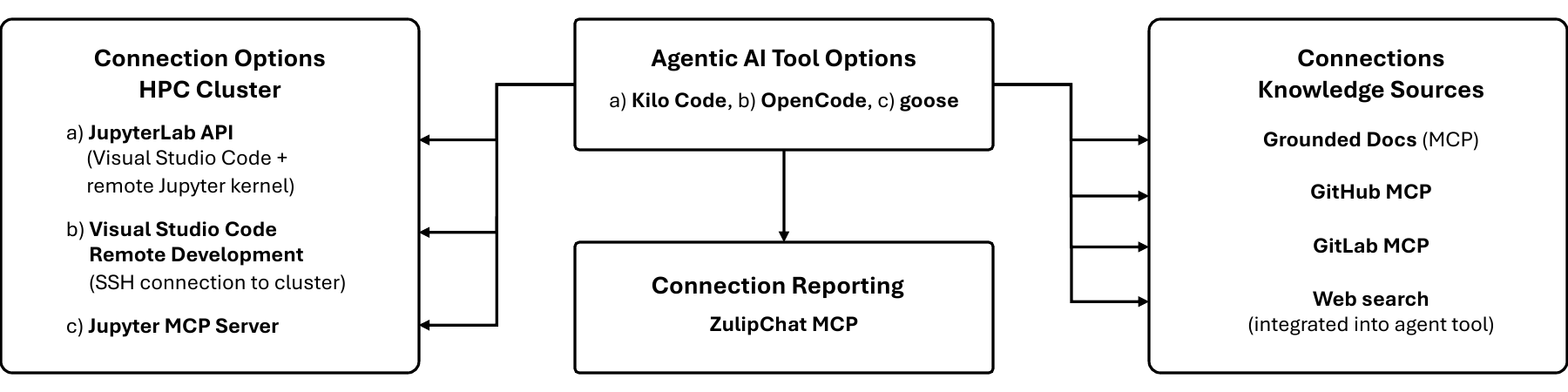}
  \caption{Agentic AI system components and recommended solutions per component}
  \label{fig:components}
\end{figure*}

The rapidly evolving \gls{ai} landscape offers many highly popular agentic \gls{ai} tools and extensions.
Based on the learnings from the evaluation of our first, internally developed prototype, we wanted to create a second, state-of-the-art prototype that could be adapted and maintained in the long term by members and users of European XFEL.
We therefore shifted our focus to the evaluation of off-the-shelf components and added minimal effort for implementation and replacement to the previously analyzed requirements (requirements 4 and 5 in \autoref{tab:req}).

We decomposed the system into four components: an agentic tool, connectivity to the \gls{hpc} cluster, connectivity to multiple knowledge sources, and reporting.
In March 2026, we evaluated a set of free, open-source tools for each component.
These tools can be executed without external telemetry to meet the privacy-oriented requirement.
\autoref{fig:components} provides an overview of the components and recommended solutions.

The most critical decision concerned the connection to the \gls{hpc} cluster.
We selected and evaluated the \gls{api} of Jupyter Lab, the Jupyter MCP Server, and Visual Studio Code Remote Development as possible solutions.
The Jupyter Lab \gls{api} was the most stable option for our environment because it connected a local notebook file to a remote Jupyter kernel of the cluster while giving an agentic tool read, write, and execution access to the file.
The Jupyter MCP Server was a good agentic solution in principle, but we could not use it in combination with the \gls{hpc} cluster because its dependencies did not meet the configuration of the cluster.
Visual Studio Code Remote Development also proved unreliable for our use case because of interrupted connections and long latencies.
Therefore, we favored the Jupyter Lab \gls{api} in combination with Visual Studio Code to connect to the cluster.

For the agentic tool, Kilo Code, OpenCode, and goose were the most promising options to us because they supported local \glspl{llm}, the \gls{mcp} for connectivity with other systems, parallel-processed sub-agent workflows, and natural-language customization through mechanisms such as \textit{AGENTS.md} and skill files.
Jupyter AI and the DeepAgents framework were less suitable.
Jupyter AI offered fewer flexible customization options and required installation on the Jupyter Lab server, which was not reliably possible in the \gls{hpc} environment.
DeepAgents would have required building a custom agentic system from scratch.
We selected the Visual Studio Code extension Kilo Code for our second prototype because it aligned best with the Jupyter Lab \gls{api} solution.

For knowledge access, we connected the agentic tool to GitHub, the European XFEL GitLab server, and the \gls{rag} system Grounded Docs.
The agent accessed these sources through \gls{mcp} servers.
GitHub was the simplest case because the official \gls{mcp} server provided a stable interface to public and private repositories.
GitLab required a third-party solution, since the official version depended on configuration changes that regular users could not make on the European XFEL server.
For knowledge retrieval, Grounded Docs was preferable to OpenRAG because it remained lightweight and quickly returned relevant documentation chunks that the agent could synthesize into its own response.
In comparison, OpenRAG is a more heavyweight, standalone system that returns an already synthesized natural-language response via its \gls{mcp} extension.
Additionally to the \gls{mcp} servers, we allowed the agent to use its integrated feature for retrieving internet search results.

Reporting agent conversations to staff members is a feature that the managers originally identified.
The agent should send reports to the Zulip messenger, which is a popular tool at European XFEL.
We selected a third-party \gls{mcp} server that enabled sending messages via a chatbot persona.

This tool evaluation yielded a second prototype built around Kilo Code, the Jupyter Lab \gls{api}, and \gls{mcp} connections to GitHub, GitLab, Grounded Docs, and Zulip.
Although we did not design the evaluation process as a quantitative benchmark study, it was systematic and tailored to our iterative \gls{dsr} approach.
The combination of the selected off-the-shelf solutions met all previously analyzed system requirements, which we demonstrate next.

\subsection{Design, Demonstration, Evaluation of Second Prototype}

\begin{figure*}[t]
  \centering
  \includegraphics[width=\textwidth]{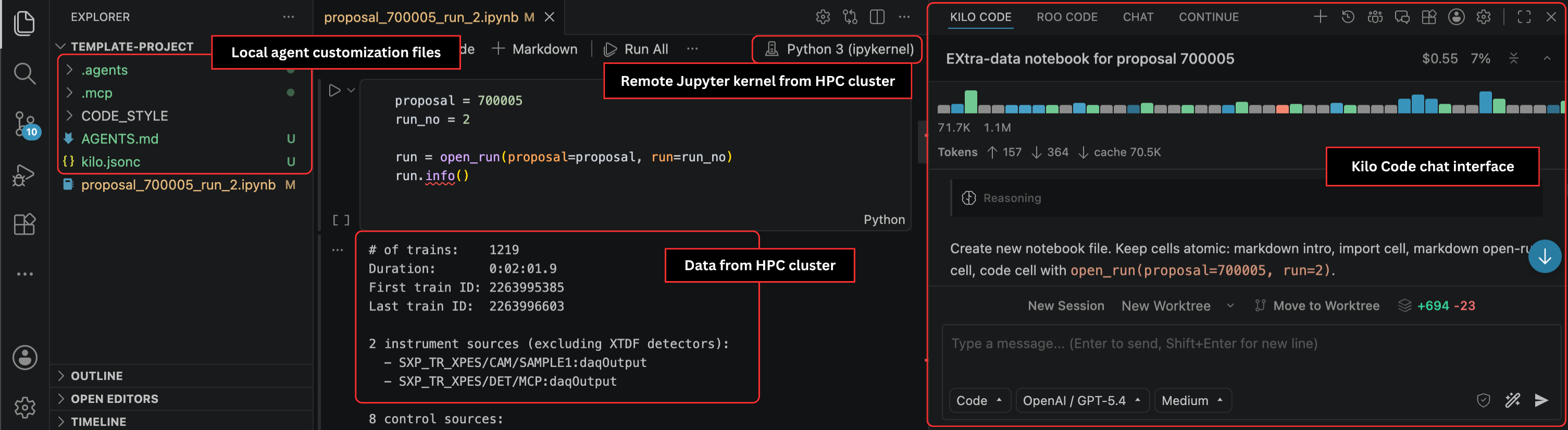}
  \caption{Data analysis in Visual Studio Code with Kilo Code agent, agent customization files, and data and Jupyter kernel from HPC cluster}
  \label{fig:prototype}
\end{figure*}

\begin{table}
  \centering
  \caption{(RQ3) design recommendations for the agentic AI system}
  \begin{tabularx}{\linewidth}{C{0.2cm} X}
    \toprule
    \textbf{No.} & \textbf{Design recommendation} \\
    \midrule
    & \textbf{System architecture:} \\
    1 & Decide on system architecture based on variance of user workflows. \\
    2 & Check user's tool expectations before recommending agentic tool. \\
    \midrule
    & \textbf{General agent behavior:} \\
    3 & Use programmatic hooks to force the agent to adhere to behavior. \\
    4 & Configure required types of human approvals in advance. \\
    \midrule
    & \textbf{Knowledge retrieval:} \\
    5 & Make available documentation agent-ready. \\
    6 & Steer agent toward relevant knowledge sources. \\
    7 & Force order of knowledge sources during knowledge retrieval. \\
    8 & Enable user to customize knowledge retrieval workflow. \\
    9 & Use small LLM with medium/high reasoning for knowledge retrieval. \\
    \midrule
    & \textbf{Code generation:} \\
    10 & Force spec-driven development. \\
    11 & Force iterative prototyping of minimal solutions. \\
    12 & Forbid hard-coding default values for mandatory user input. \\
    13 & Force request to clarify the execution environment. \\
    14 & Use large LLM with low/medium reasoning for code generation. \\
    \bottomrule
  \end{tabularx}
  \label{tab:recom}
\end{table}

Based on the requirements analysis and components evaluation, we designed a second prototype based on the selected off-the-shelf tools (\autoref{fig:prototype}), demonstrated its applicability for offline data analysis, and evaluated and iteratively improved the prototype during a user study in April 2026.
Afterward, we derived 14 design recommendations for maintainable \gls{ai} support (\autoref{tab:recom}).

A major part of the design process was the adjustment of agent behavior.
We used an AGENTS.md file to explain the prepared code project, the expected data analysis workflow, and the behavioral rules to the agent.
The rules corresponded to the system requirements 6 to 19 (\autoref{tab:req}).
We added requirement 10 (\textit{use concise language}) after encountering verbose responses from the agent during initial experiments.
To transform these responses into a concise, action-oriented list of bullet points, we instructed the agent to use the \textit{caveman} skill (in lite mode), which was a popular skill at that time~\cite{brussee_caveman_2026}.

We evaluated the prototype in 13 one-hour sessions between the first author and individual European XFEL members.
These included eight members with an occupation in data science or computer science, as well as five with an occupation in photon science.
Participants brought their own unique use cases, ranging from visualizing data and analyzing specific data aspects to preparing batch jobs for the \gls{hpc} cluster.

Participants appreciated the agent behavior and responsiveness, the quality of knowledge retrieval results and generated source code, and the ability to execute Jupyter notebooks on the \gls{hpc} cluster.
Furthermore, they considered the reporting feature useful for accelerating support processes.

\begin{lstlisting}[
   float=tbp,
   caption={AGENTS.md instructions for code execution on HPC cluster},
   label={lst:agentsmd-test-code}
]
## Execute and Test Code on HPC Cluster

Always follow these steps after you generated code:

1. DO NOT test the code by executing it in your
   internal sandbox. Instead, create a new code
   file or edit an existing one (e.g., a Jupyter
   notebook) and execute the file with an available
   Jupyter kernel or Python environment.

   - ALWAYS ask the user first, which Jupyter kernel
     or Python environment you should use to execute
     the code.
   - You MUST NOT execute the code in your internal
     sandbox as the sandbox does not have access to
     the data and computational environment of the
     HPC cluster.

2. Review if any errors appeared (e.g., error
   messages in notebook cell output).

3. If errors appeared, rework your code and start
   again with step 1.
\end{lstlisting}

The evaluation also exposed recurring friction points.
We had to force the agent to execute, test, and refine code directly within the notebook rather than in its internal sandbox, so that it could use the data, libraries, and computational resources of the cluster (\autoref{lst:agentsmd-test-code}).
When we used small \glspl{llm} for chat completions, the agent sometimes failed to reuse the code of European XFEL libraries.
Additionally, the system also produced many explicit approval requests during the first sessions, including for uncritical read actions.
These observations led us to refine the prototype iteratively.

We updated the prototype during and after each session.
Typical updates during a session involved adding documentation to Grounded Docs to accommodate the specific use cases of participants.
After the sessions, we updated the instructions in the AGENTS.md file if the agent did not behave as the participant or we had expected.
Overall, we found natural-language instructions not sufficient to ensure stable behavior.
The agent reacted differently depending on the selected \gls{llm}, reasoning level, and wording.
Programmable \textit{hooks} would constitute more reliable options for instructions.
At that time, however, Kilo Code did not yet support these hooks.
Instead, it supported explicit rule files, which we tested during one session.
However, they were harder to maintain and less interoperable than an AGENTS.md file, and did not lead to improved agent behavior.

The 14 design recommendations emphasize, for example, the need for configurable system components tailored to user needs, explicit human approval for (exclusively) sensitive actions, programmable hooks to complement natural-language instructions, and agent-compatible documentation.
Overall, the second prototype demonstrated that a selection of off-the-shelf tools can be adapted to the offline data analysis workflows at European XFEL.
However, this was possible only through continuous design iterations.

\subsection{Answers to the Research Questions}
\label{sec:answers}

Our answers to the research questions of this study characterize the challenges of offline data analysis at European XFEL, as well as the requirements and design recommendations for an agentic \gls{ai} system that addresses the challenges.

\textbf{Answering RQ1,} we found 13 partially overlapping knowledge challenges of facility users, Data Analysis group members, and staff scientists during offline data analysis at European XFEL (\autoref{tab:chal}).
The key challenge is that users must turn scattered and partially tacit knowledge into a project-specific data analysis workflow.
The difficulty is not only clarifying analysis objectives and understanding the complex infrastructure and software, but also synthesizing various documentation resources and informal support.
Therefore, the fundamental problem in practice is the integration and contextualization of knowledge~\cite{robillard_-demand_2017}, and not necessarily the production of documentation.

\textbf{Answering RQ2,} we analyzed 21 requirements that translate the challenges into concrete design objectives for an agentic \gls{ai} system (\autoref{tab:req}).
A useful system must be integrated with the \gls{hpc} cluster, support both knowledge retrieval and code generation, provide verified citations and code testing, and enforce human oversight for sensitive actions.
The additional requirements for minimal implementation and replacement effort are especially important for this study because they capture the need for a maintainable system that can adapt to an evolving tool landscape.

\textbf{Answering RQ3,} we derived 14 design recommendations that show how the requirements can be implemented in practice (\autoref{tab:recom}).
They suggest that maintainable \gls{ai} support depends on the right balance between flexibility and control: off-the-shelf tools are preferable to custom-built systems, but only when they can be configured to match required user workflows and safety constraints.
The recommendations also reinforce two central lessons of the study: documentation must be agent-ready, and behavioral control should rely on programmable hooks rather than on natural-language instructions alone.

Overall, the answers to the research questions support the main objective of the study: to move scientists from overload to insights by designing an agentic \gls{ai} system that can integrate knowledge retrieval, code generation, and the \gls{hpc} environment in a maintainable way.
The findings show that this is possible, but only if the system is grounded in the user workflow, controlled carefully, and built from components that can evolve with the surrounding tool ecosystem.

\section{Discussion}
\label{sec:disc}

\subsection{AI Agents Become Standard Workflow Components}

Our findings suggest that \gls{ai} support is becoming a normal part of the workflows at European XFEL.
This trend becomes particularly evident when we compare the current \gls{ai} usage of the user study participants with an earlier study conducted at the facility~\cite{kruse_can_2024}.
What started as occasional assistance for knowledge retrieval is increasingly moving toward agents that are embedded in everyday coding and analysis work.

Available tools are already sufficient to support different user needs and preferences.
Our tool analysis showed that the combination of a local agentic tool and \gls{mcp} servers, including a lightweight \gls{rag} service, can cover the requirements of offline data analysis.
A working system can be configured by a facility user without changes to the existing infrastructure and software, which makes practical adoption feasible.

At the same time, broader adoption of agentic tools at European XFEL still depends on careful testing, clear guidance, and management support.
Beyond technical feasibility, the remaining challenges are safe operation, robustness, and helping users configure the tools in ways that fit their workflows.
In that sense, \gls{ai} agents should be treated less as add-ons and more as emerging, soon indispensable workflow components that need deliberate integration into the facility environment.

\subsection{Maintainable AI Systems Require Flexible Components}

Our analysis suggests that maintainability depends less on reducing the number of system parts than on keeping those parts modular and replaceable.
The second prototype was more complex than the first one, but it was also easier to maintain because the agentic core, the knowledge source connectors, and the reporting mechanism could be used out of the box and configured independently.
In the European XFEL setting, a more elaborate configuration can be more sustainable than a self-developed solution if each component can be swapped or adjusted without rebuilding the whole system.

The tool evaluation showed that several technically feasible combinations of components are available, but only a few are likely to fit everyday work at European XFEL.
Users could connect to the \gls{hpc} cluster through local or remote Visual Studio Code sessions, the browser-based Jupyter Lab, or programmatic \gls{mcp} interfaces.
Furthermore, some agentic tools expose multiple frontends, including command-line interfaces, standalone desktop apps, and web apps.
Theoretically, users could use the OpenCode web interface to orchestrate the data analysis process from their smartphones.
In practice, however, the most relevant options are a local Visual Studio Code window, connected to the Jupyter Lab \gls{api} and Kilo Code, or a similar setup based on a remote SSH connection.

Another maintainability aspect concerns behavioral control.
Our user study showed that AGENTS.md and skill files are useful for describing workflows, but they do not produce stable behavior across different models, reasoning levels, and instruction formulations.
Reliable workflows, therefore, need programmable hooks rather than only natural-language instructions, especially when tasks such as code commits or other traceable actions should happen reliably.
As tools such as OpenCode and Claude Code already support hooks, we expect that this will become a standard feature soon.
At the same time, natural-language configuration options remain valuable for rules that are easier to express and maintain in prose.
Because the landscape of \gls{ai} tools continues to evolve rapidly, a maintainable agentic \gls{ai} system should apply interoperable configuration options that enable a low-effort transition to new system components and protocols.

\subsection{Agents Require AI-First Documentation}

If European XFEL users and members increasingly rely on agents to retrieve knowledge, future-proof technical documentation must be designed for \gls{ai} consumption.
That means the documentation must be structured, complete, and available in lightweight formats that agents can process reliably.

Our user study suggests that \gls{api} reference documentation alone is not sufficient.
Participants often valued the generated code, but they also observed that the agent missed library-specific details when the available documentation was incomplete or too superficial.
This matches the move of the Data Analysis group toward cookbook-style guidelines for selected scientific instruments and experiment types, because workflow-focused examples give humans and \gls{ai} agents the context they need to produce correct data analysis code.

The paradigm shift toward \gls{ai}-first documentation resembles mobile-first web design, in which smartphone screens determine the structure of modern websites.
For documentation, that implies maintaining an agent-consumable format alongside or instead of the existing resources.
One practical option for websites is adding a Markdown-based \textit{llms.txt} file~\cite{howard_llmstxt_2024}.
Such a layer can reduce parsing problems and lower the retrieval overhead of web scrapers.
Its value, however, depends on consistent maintenance by documentation owners and on scrapers that actually focus on the simplified format~\cite{mueller_crawlers_2026}.

\subsection{Ethical Aspects Remain Essential for AI Acceptance}

Multiple research activities of this study showed that ethical aspects are essential for the acceptance of \gls{ai} support in complex scientific environments.
In our setting, the primary concerns centered on traceability of retrieved information, transparency of internal agent workflows, and human oversight regarding sensitive actions.
Study participants were willing to rely on the agent when it cited its sources, explained what it was doing, and left room for user intervention.

Citations played an important role in building trust during knowledge retrieval.
Participants could trace the sources of the response through the linked references in the agent output and check whether the cited material supported the recommendation.
This made the retrieval process more inspectable and helped distinguish grounded answers from unsupported ones.

Implementation plans were equally important because they made the agentic reasoning process easier to review.
By showing a checklist of steps before execution, the system allowed users to inspect the planned behavior and adjust it if needed.
This was especially relevant in our scientific setting, where the generated code directly affects the results of scientific analyses.

Human approval remained necessary for actions that interact with the real \gls{hpc} environment.
The prototype had to execute and test code against actual data, libraries, and compute resources.
This creates risks if the agent would act without oversight.
The explicit approval requests had to focus on code edits and execution rather than on read operations to ensure that human attention was directed toward critical decisions.

These observations suggest that ethical guidance for agentic systems should not remain in theoretical guidelines.
A promising research approach involves adapting available taxonomies of ethical \gls{ai} requirements~\cite{puhlfurs_model_2025} to agentic systems and executing recurring ethical reviews using an agentic skill file.
Such a skill would bring ethical requirements closer to modern \gls{ai} workflows and facilitate their implementation.

\section{Limitations}
\label{sec:lim}

We discuss the study design trade-offs and the threats to validity in the methodology section, as recommended by \citet{robillard_communicating_2024}.
The current section covers the remaining limitations specific to the study objectives.

The study is centered on offline data analysis at European XFEL.
The agentic \gls{ai} system, including all customization options, is tailored to this task and the facility infrastructure.
Transferability to other tasks at European XFEL and other scientific facilities is uncertain.
We also expect facility users to further tailor the AGENTS.md file to their individual needs.
Offline data analysis encompasses a wide range of use cases, and users possess varying levels of expertise.
Hence, users have to adjust the agent to their needs, including the level of computational optimization required for their use case.
Nevertheless, practitioners at and outside of European XFEL can compare the analyzed challenges and system requirements with those in their environment and adapt our system design recommendations and project artifacts accordingly.

Only European XFEL staff tested the prototypes, not actual users.
We chose staff members as proxies for real facility users due to organizational restrictions.
The selected staff members had experience with the data analysis process, but to a varying degree.
This reflects the data analysis skills of actual users, ranging from unfamiliar to regular visitors.
For example, the user study included sessions with two PhD students who were the closest proxy for novice facility users.
The students work on specific research projects requiring beamtime, contribute to beamtime preparation, and are typically responsible for the associated data analysis.

The two managers we interviewed in the initial phase of the study became co-authors of this paper because they supported the design of the subsequent research activities.
Both also took part in the user study as participants because they were not actively involved in the prototype development.
During the user study, they focused on their roles as managers to provide honest feedback about the prototype.

We have not tested production-grade aspects of the prototype in detail.
For example, a malicious user could use the system to flood the \gls{hpc} cluster with requests.
In practice, however, we consider this security aspect as a minor threat with respect to the proposed system architecture.
The agent has the same permissions and restrictions as the user who configured the system.
Anything the agent can do, a user could do as well.
What we have to test further, however, are safety aspects to prevent users from unintentionally causing harm through an agent acting autonomously.
A possible solution is a sandbox that encloses the agent and grants access only to specific server directories and capabilities.

The tool comparison is not a benchmark study, and we did not evaluate the prototype during a long-term deployment.
Selected system components, such as Kilo Code and Grounded Docs, may not be the best options for our use case.
However, they were good enough as they fulfilled the previously established requirements.
Practitioners of other organizations can choose a different system configuration.
However, as many agentic tools and \gls{mcp} servers are interoperable, they can still reuse our design recommendations.
A long-term deployment of the system was outside the scope of this exploratory study and is reserved for future work.

\section{Related Work}
\label{sec:rw}

Recent work on \gls{ai} support for \textbf{scientific data analysis} suggests that assistants should combine planning and execution steps, align with scientific workflows, and adapt to a user's expertise.
\citet{gu_how_2024} show that effective scientific \gls{ai} assistants combine coding with planning support, such as providing suggestions, alternatives, and rationale.
In their study, they find that explanations of an assistant can build trust among scientists but become distracting when they are too verbose, pointing to the value of concise and adaptable responses.
\citet{mcnutt_design_2023} argue that assistants for computational notebooks can increase productivity while supporting the verification and reproducibility demands of scientists.
An assistant should operate across multimodal notebook context, support selective code execution, offer domain-specific linters, and have accessible control features.
\citet{ramasamy_ai_2025} find that code recommendations of workflow-aware assistants can have high acceptance, especially for predictive scientific workflows.
\citet{cristea_jelai_2025} demonstrate how context-aware tutoring can be embedded directly in notebooks.
Together, these studies suggest that effective scientific assistants are tightly integrated into the analysis environment to provide grounded knowledge.

Research on \gls{ai} assistants for various \textbf{coding tasks} shows that they can accelerate routine programming tasks, but their usefulness declines when project context, external dependencies, or domain-specific constraints matter.
In a previous study, we found that predefined prompts can produce more consistent and readable code documentation than ad-hoc prompts, but that experienced developers like the interactiveness of ad-hoc prompting~\cite{kruse_can_2024}.
\citet{corso_generating_2024} show that code assistants often fail or hallucinate when dependencies and broader project context become important.
\citet{nettur_role_2025} summarize that coding assistants provide productivity gains but also introduce security, reproducibility, and intellectual-property risks.
Based on that, \citet{murali_ai-assisted_2024} demonstrate that domain-adapted assistants can be adopted at scale in production when they are integrated into the environment with safeguards.
Overall, the literature indicates that practitioners must evaluate code assistants for context awareness, validation support, and safe integration into development workflows.

Further work on the \textbf{risks of \gls{ai} support} emphasizes human control, traceability, and verifiability as mitigation strategies.
\citet{rahe_how_2025} show that inexperienced developers can become overreliant on generated code and stop evaluating it critically.
\citet{tang_risks_2025} argue for a safety-first approach with constrained autonomy, traceability, and expert oversight.
\citet{kazemitabaar_improving_2024} find that providing an editable, data-grounded implementation plan improves scientists' ability to verify assistant behavior.
In an earlier study, we derived ethical requirements that call for stronger \gls{ai} documentation and tooling to operationalize and validate risk mitigation~\cite{puhlfurs_model_2025}.
These studies point to the need for transparent, bounded, and reviewable AI support in scientific settings.

Our study builds on these strands by bringing them together in a facility-specific \gls{dsr} project.
Instead of examining notebook assistance, code generation, or risk mitigation in isolation, we study and report how an agentic system can support offline data analysis at European XFEL for knowledge retrieval and code generation while working on a real \gls{hpc} cluster.
This lets us translate prior findings into concrete requirements and design recommendations for a maintainable, human-controlled system that fits a specialized infrastructure.

\section{Conclusion}
\label{sec:conc}

This work addresses the problem of offline data analysis at large scientific facilities like European XFEL.
This includes complex data, fragmented knowledge, and heterogeneous user expertise, which makes effective support hard to scale.
Using a \gls{dsr} approach, we iteratively analyzed knowledge challenges, derived software requirements, built two prototypes, and evaluated them with facility experts.

We identified 13 challenges such as users' unfamiliarity with the infrastructure, software, and data analysis procedures, as well as the overall time required for staff members to provide support.
The key challenge for users is the need to synthesize scattered documentation, informal support, and data analysis expertise into a project-specific data analysis workflow.

We analyzed 21 requirements showing that a useful \gls{ai} agent must be integrated into the \gls{hpc} environment, grounded in retrievable knowledge, able to generate and test source code, and constrained by human oversight for sensitive actions.
A critical requirement is that system components must remain replaceable and low-maintenance so the system can adapt to user needs and the evolving \gls{ai} landscape.

Our user study suggests 14 design recommendations for an agentic \gls{ai} system.
We showed that off-the-shelf agent tools are more practical than a custom-built system, but only when they can be configured to match user workflows and safety constraints.
In practice, that means documentation must be agent-ready, approvals must be configured deliberately, and behavioral control should rely on programmable hooks rather than on natural-language instructions alone.

Future research should test these recommendations in longer deployments and in additional scientific workflows to determine which parts generalize beyond offline data analysis at European XFEL.
In particular, programmable hooks that could ensure consistent agent behavior deserve further study.
Scientific facilities will support such agentic systems in their critical environments only if reliable guardrails are in place.

Overall, the study shows that an agentic \gls{ai} system can help move scientists from overload to insights by connecting knowledge retrieval, code generation, and the \gls{hpc} environment into an integrated, modular system.
This promise depends on careful grounding, explicit safeguards, and continuous adaptation.
Without these constraints, the same system would amplify noise rather than reduce it.

\section*{Acknowledgment}
\label{sec:ack}

We thank all participants for their insights, in particular the Data Analysis group of European XFEL.
We acknowledge DASHH, Data Science in Hamburg - Helmholtz Graduate School for the Structure of Matter, for financial support.

\bibliographystyle{IEEEtranN}
\bibliography{bib}

\end{document}